\documentclass[11pt,a4paper]{article}
\usepackage[hyperref]{emnlp2020}
\usepackage{times}
\usepackage{latexsym}

\usepackage{url}

\usepackage{graphicx}
\usepackage{CJKutf8}
\usepackage{multirow}
\usepackage{array}
\usepackage{tabularx}
\usepackage{makecell}
\usepackage{amsmath} 
\usepackage{amssymb}
\usepackage{tablefootnote}
\usepackage{subcaption}
\usepackage{diagbox}
\usepackage{graphicx}
\usepackage{algorithm}
\usepackage[noend]{algpseudocode}
\usepackage{booktabs}
\usepackage{xcolor}
\usepackage{color, soul} 
\usepackage{colortbl}
\usepackage{makecell}
\definecolor{orange}{rgb}{1,0.5,0}

\usepackage{microtype}

\newcolumntype{L}[1]{>{\raggedright\arraybackslash}p{#1}}
\newcolumntype{C}[1]{>{\centering\arraybackslash}p{#1}}
\newcolumntype{R}[1]{>{\raggedleft\arraybackslash}p{#1}}

\usepackage{microtype}

\aclfinalcopy 
\def\aclpaperid{3164} 

\newcolumntype{L}[1]{>{\raggedright\arraybackslash}p{#1}}
\newcolumntype{C}[1]{>{\centering\arraybackslash}p{#1}}
\newcolumntype{R}[1]{>{\raggedleft\arraybackslash}p{#1}}

\DeclareSymbolFont{extraup}{U}{zavm}{m}{n}
\DeclareMathSymbol{\varheart}{\mathalpha}{extraup}{86}
\DeclareMathSymbol{\vardiamond}{\mathalpha}{extraup}{87}

\title{Improving Named Entity Recognition with \\Attentive Ensemble of Syntactic Information}

\author{
    Yuyang Nie$^{\diamondsuit*}$, \hspace{0.1cm}
    Yuanhe Tian$^{\varheart*}$, \hspace{0.1cm}
    Yan Song$^{\spadesuit\heartsuit\dag}$, \hspace{0.1cm}
    Xiang Ao$^{\triangle\Box}$, \hspace{0.1cm}
    Xiang Wan$^{\heartsuit}$ \\
    $^{\diamondsuit}$University of Electronic Science and Technology of China \\
    $^{\varheart}$University of Washington \hspace{0.2cm} 
    $^{\spadesuit}$The Chinese University of Hong Kong (Shenzhen)\\
    $^{\heartsuit}$Shenzhen Research Institute of Big Data\\
    $^{\triangle}$Institute of Computing Technology Chinese Academy of Sciences\\
    $^{\Box}$University of Chinese Academy of Sciences\\
    \tt  $^{\diamondsuit}$nyy207@gmail.com $^{\varheart}$yhtian@uw.edu \\
    \tt $^{\spadesuit}$songyan@cuhk.edu.cn
    $^{\triangle}$\texttt{aoxiang@ict.ac.cn}
    $^{\heartsuit}$wanxiang@sribd.cn
}

\date{}

\begin{document}
\maketitle

\def\thefootnote{*}\footnotetext{Equal contribution.}
\def\thefootnote{\dag}\footnotetext{Corresponding author.}

\def\thefootnote{\arabic{footnote}}

\begin{abstract}
Named entity recognition (NER) is highly sensitive to sentential syntactic and semantic properties where entities may be extracted according to how they are used and placed in the running text.
To model such properties, one could rely on existing resources to providing helpful knowledge to the NER task; some existing studies proved the effectiveness of doing so, and yet are limited in appropriately leveraging the knowledge such as distinguishing the important ones for particular context.
In this paper,
we improve NER by leveraging different types of syntactic information through
attentive ensemble,
which functionalizes by the proposed
key-value memory networks, syntax attention, and the gate mechanism for encoding, weighting and aggregating such syntactic information, respectively.
Experimental results on six English and Chinese benchmark datasets suggest the effectiveness of the proposed model
and show that it outperforms previous studies on all experiment datasets.\footnote{The code and the best performing models are available at \url{https://github.com/cuhksz-nlp/AESINER}}
\end{abstract}

\section{Introduction}
\label{intro}



Named entity recognition (NER) is one of the most important and fundamental tasks in natural language processing (NLP), 
which identifies named entities (NEs), such as locations, organizations, person names, etc., in running texts,
and thus
plays an important role in downstream NLP applications including question answering \cite{DBLP:conf/aaai/PangLGXSC19}, semantic parsing \cite{DBLP:conf/acl/LapataD18} and entity linking \cite{DBLP:conf/acl/MartinsMM19}, etc.
%
%

The main methodology for NER is conventionally regarded as a sequence labeling task with models such as hidden Markov model (HMM) \cite{DBLP:conf/anlp/BikelMSW97} and conditional random field (CRF) \cite{DBLP:conf/conll/McCallum003} applied to it in previous studies.
Recently, neural models play a dominate role in this task and illustrated promising results \cite{DBLP:journals/jmlr/CollobertWBKKK11,DBLP:journals/corr/HuangXY15,DBLP:conf/naacl/LampleBSKD16,DBLP:conf/emnlp/StrubellVBM17,DBLP:conf/coling/YadavB18,DBLP:conf/aaai/ChenLDLZK19,DBLP:conf/emnlp/JieL19,DBLP:conf/acl/LiuMZXCZ19,DBLP:conf/emnlp/BaevskiELZA19}, 
because they are powerful in encoding contextual information and thus drive NER systems to better understand the text and recognize NEs in the input text.
%
%
\begin{figure*}[t]
    \centering
    \includegraphics[width=\textwidth, trim=0 30 0 0]{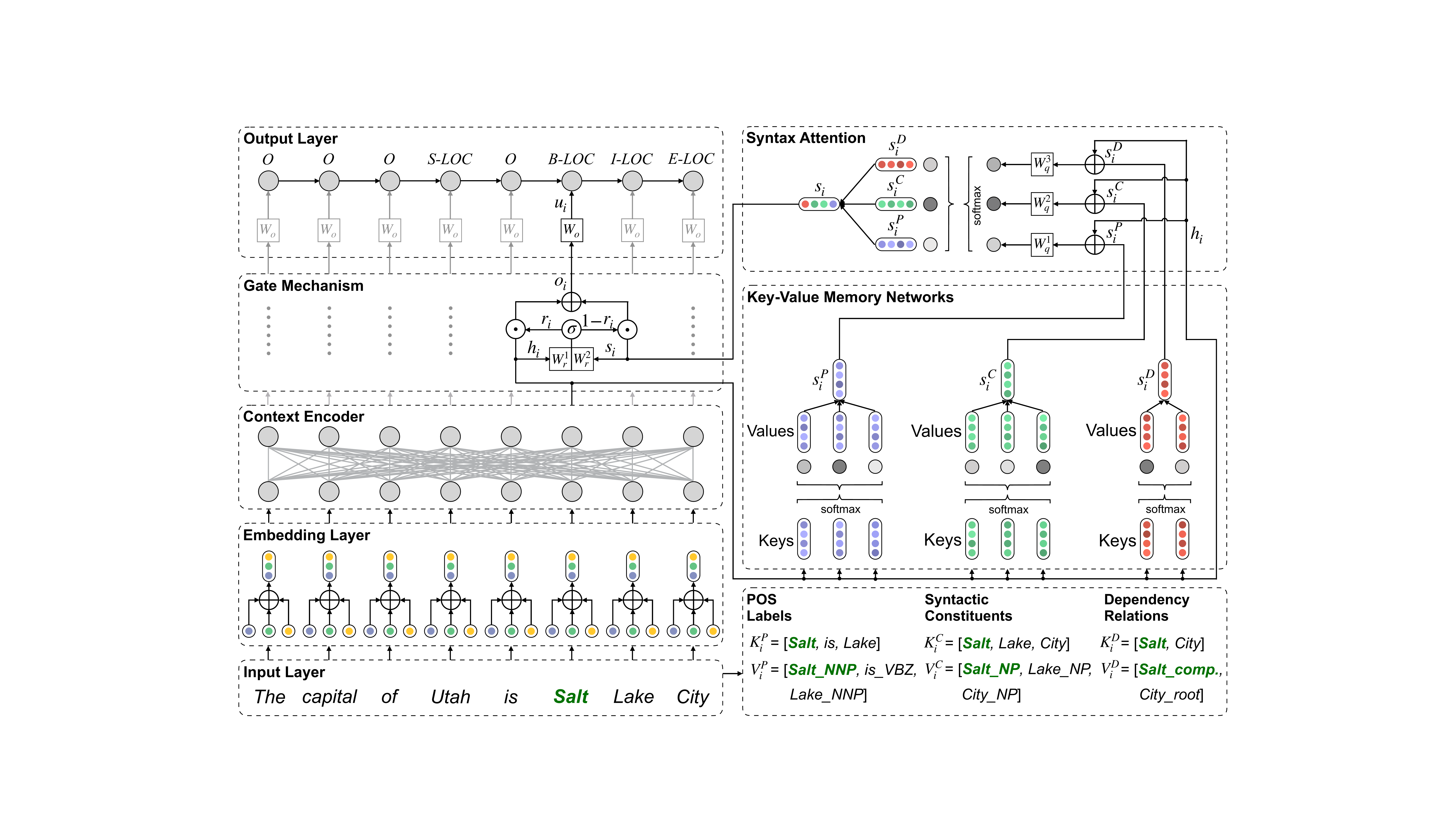}
    \caption{The overall architecture of the proposed NER model integrated with attentive ensemble of different syntactic information.
    An example input sentence and its output entity labels are given and the 
    syntactic information for the word ``\textit{Salt}'' are illustrated with their processing through KVMN, syntax attention and the gate mechanism.
    %
    }
    \label{fig:model}
    \vskip -1em
\end{figure*}
%
%
%
%
%
%
%
%
Although it is straightforward and effective to use neural models
to help NER, it is expected to incorporate more useful features into an NER system.
Among all such features, syntactic ones, such as part-of-speech (POS) labels, syntactic constituents, dependency relations, are of high importance to NER because they are effective in identifying the inherited structure in a piece of text and thus guide the system to find appropriate NEs accordingly, which is proved in a large body of previous studies \cite{DBLP:conf/uai/McCallum03,DBLP:conf/emnlp/LiDWCM17,DBLP:journals/bioinformatics/LuoYYZWLW18,DBLP:journals/bioinformatics/DangLNV18,DBLP:conf/emnlp/JieL19}.
Although promising results are obtained, existing models are limited in regarding extra features as gold references and directly concatenate them with word embeddings.
Therefore,
such features are not distinguished and separately treated when they are used in those NER models, where the noise in the extra features (e.g., inaccurate POS tagging results) may hurt model performance.
As a result,
%
it is still a challenge to find an appropriate way to incorporate external information into neural models for NER.
%
Moreover, in most cases, one would like to incorporate more than one types of extra features. 
Consequently, it is essential to design an effective mechanism to combine and weight those features so as to restrict the influence of noisy information.

In this paper, we propose a sequence labeling based neural model to enhance NER by incorporating different types of
syntactic information,
which is conducted by
attentive ensemble with key-value memory networks (KVMN) \cite{miller2016key}, syntax attention and the gate mechanism.
Particularly, the KVMN is applied to encode the context features and their syntax information from different types, e.g., POS labels, syntactic constituents, or dependency relations;
syntax attention is proposed to weight different types of such syntactic information, and the gate mechanism controls the contribution of the results from the context encoding and the syntax attention to the NER process.
%
%
Through the attentive ensemble, important syntactic information is highlighted and emphasized during labeling NEs.
%
%
In addition, to further improve NER performance, we also try different types of pre-trained word embeddings, which is demonstrated to be effective in previous studies \cite{DBLP:conf/coling/AkbikBV18,DBLP:conf/emnlp/JieL19,DBLP:conf/acl/LiuYL19,DBLP:journals/corr/abs-1911-04474}.
We experiment our approach on six widely used benchmark datasets from the general domain, where half of them are in English and the other half are in Chinese.
Experimental results on all datasets suggest the effectiveness of our approach 
to enhance NER through syntactic information, where state-of-the-art results are achieved on all datasets.


\begin{figure*}[t]
    \centering
    \includegraphics[width=\textwidth, trim=0 30 0 0]{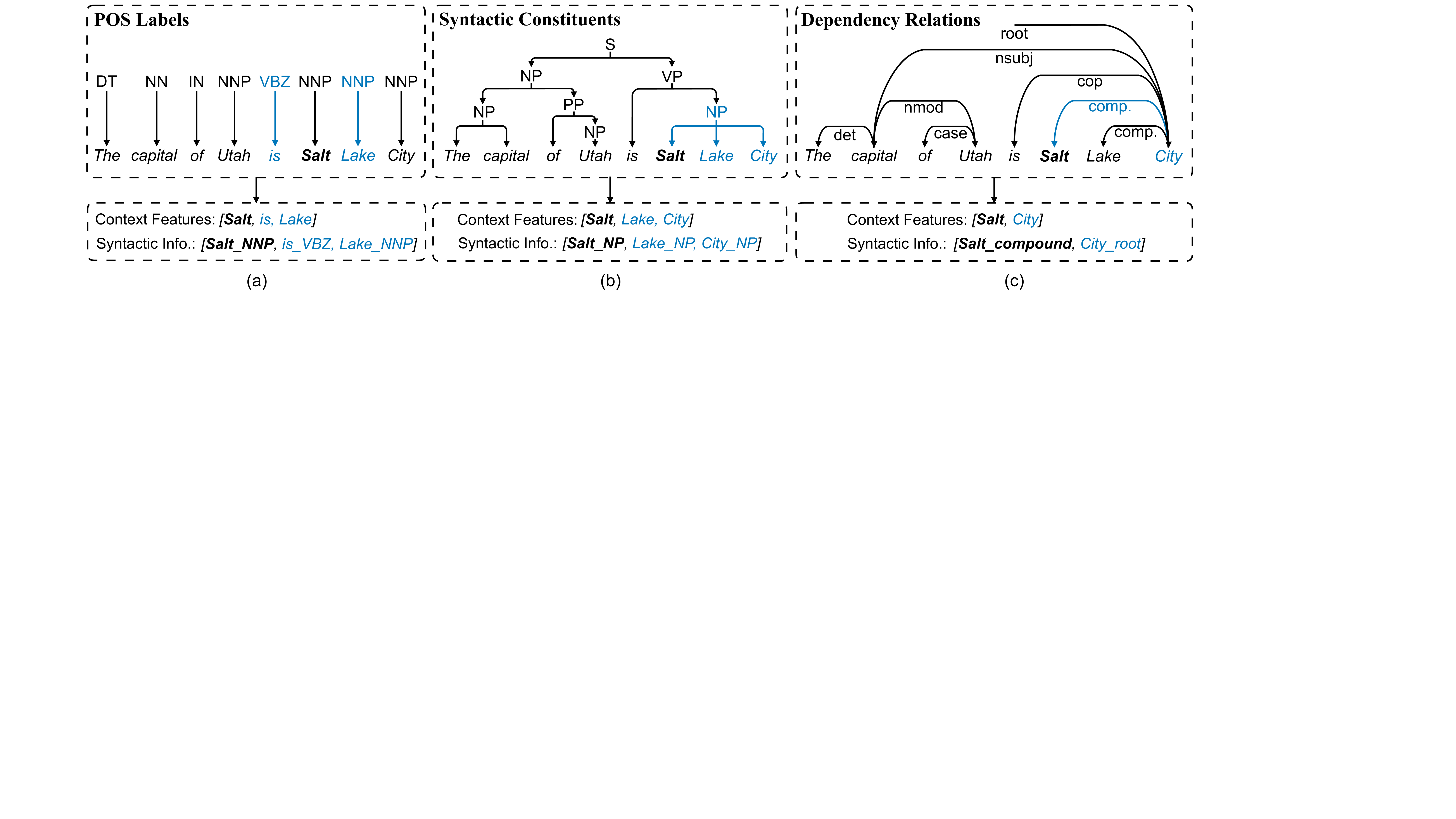}
    \caption{
    The extracted syntactic information in POS labels (a), syntactic constituents (b), and dependency relations (c) for \textit{``Salt''} in the example sentence, where associated contextual features and the corresponding instances of syntactic information are highlighted in blue.
    }
    \label{fig:knowledge}
    \vskip -1em
\end{figure*}


\section{The Proposed Model}
\label{model}



NER is conventionally regarded as a typical
sequence labeling task, where an input sequence $\mathcal{X} = x_1, x_2, \cdots, x_{i}, \cdots, x_n$ with $n$ tokens is annotated with its corresponding NE labels $\widehat{\mathcal{Y}} = \widehat{y}_1, \widehat{y}_2, \cdots, \widehat{y}_{i}, \cdots, \widehat{y}_n$ in the same length.
Following this paradigm, we propose a neural NER model depicted in Figure \ref{fig:model}
with attentive ensemble to incorporate different types of
syntactic information,
%
%
%
%
where it can be conceptually formalized by
\begin{equation}
    \widehat{\mathcal{Y}} = f(\mathcal{GM}(\mathcal{X}, \mathcal{SA}([\mathcal{M}^{c}(\mathcal{K}^{c}, \mathcal{V}^{c})]_{c \in \mathcal{C}})))
\end{equation}
where $\mathcal{C}$ denotes the set of all syntactic information types and $c$ is one of them;
$\mathcal{M}^{c}$ is the KVMN for encoding syntactic information of type $c$ with $\mathcal{K}^{c}$ and $\mathcal{V}^{c}$ referring to the keys and values in it,
respectively;
$\mathcal{SA}$ denotes the syntax attention to weight different types of syntactic information obtained through $\mathcal{M}^{c}$;
$\mathcal{GM}$ refers to the gate mechanism to control how to use the encodings from context encoder and that from $\mathcal{SA}$.
In the following text, we firstly introduce how we extract different types of syntactic information, then illustrate the attentive ensemble of syntactic information through KVMN, syntax attention, and gate mechanism,
finally elaborate the encoding and decoding of the input text for NER as shown in the left part of Figure \ref{fig:model}.

\subsection{Syntactic Information Extraction}



A good representation of the input text is the key to obtain good model performance for many NLP tasks \cite{song2017learning,sileo-etal-2019-mining}.
Normally, a straightforward way to improve model performance is to enhance text representation by embeddings of extra features, which is demonstrated to be useful across tasks \cite{marcheggiani-titov-2017-encoding,song-etal-2018-joint,zhang-etal-2019-incorporating,huang2019syntax,tian-etal-2020-constituency}, including NER \cite{DBLP:conf/acl/ZhangY18,DBLP:conf/acl/SeylerDCHW18,DBLP:conf/emnlp/SuiCLZL19,DBLP:conf/emnlp/GuiZZPFWH19,DBLP:conf/ijcai/GuiM0ZJH19,DBLP:conf/acl/LiuYL19}.
Among different types of extra features, the syntactic one
%
has been proved to be helpful in previous studies for NER, where the effectiveness of POS labels, syntactic constituents, and dependency relations, are demonstrated by \newcite{DBLP:conf/uai/McCallum03}, \newcite{DBLP:conf/emnlp/LiDWCM17}, and \newcite{DBLP:conf/tlt/CetoliBOS18}, respectively.
%
In this paper, we also focus on these three types of syntactic information.
%
In doing so, we obtain the POS labels, the syntax tree and the dependency parsing results from an off-the-shelf NLP toolkit (e.g., Stanford Parser) for each input sequence $\mathcal{X}$. 
%
Then, for each token $x_{i}$ in $\mathcal{X}$, we extract its context features and related syntactic information according to the following procedures.

\vspace{0.2cm}
For \textbf{POS labels}, we treat every $x_i$ as the central word and employ a window of $\pm1$ word to extract its context words and their corresponding POS labels.
For example, in the example in Figure \ref{fig:knowledge}(a), for ``\textit{Salt}'', the $\pm1$ word window covers its left and right words, so that the resulting context features are ``\textit{Salt}'', ``\textit{is}'', and ``\textit{Lake}'', and we use the combination of such words and their POS labels as the POS information (i.e., ``\textit{Salt\_NNP}'', ``\textit{is\_BVZ}'', and ``\textit{Lake\_NNP}'') for the NER task.

\vspace{0.2cm}

For \textbf{syntactic constituents}, we start with $x_{i}$ at the leaf of $\mathcal{X}$'s syntax tree, then search up through the tree to find the first acceptable syntactic node\footnote{There are 10 accepted constituent nodes, including $\textit{NP}$, $\textit{VP}$, $\textit{PP}$, $\textit{ADVP}$, $\textit{SBAR}$, $\textit{ADJP}$, $\textit{PRT}$, $\textit{INTJ}$, $\textit{CONJP}$ and $\textit{LST}$, which are selected from the types used in the CoNLL-2003 shared task \cite{DBLP:conf/conll/SangM03}.}, and select all tokens under that node as the context features and the combination of tokens and their syntactic nodes as the constituent information.
For example, in Figure \ref{fig:knowledge}(b), we start from ``\textit{Salt}'' and extract its first accepted node ``\textit{NP}'', then collect the tokens under ``\textit{NP}'' as the context features (i.e., ``\textit{Salt}'', ``\textit{Lake}'', and ``\textit{City}'') and combine them with ``\textit{NP}'' to get the constituent information (i.e., ``\textit{Salt\_{NP}}'', ``\textit{Lake\_{NP}}'', and ``\textit{City\_{NP}}'').

\vspace{0.2cm}

For \textbf{dependency relations}, we find all context features for each $x_{i}$ by collecting all its dependents and governor from $\mathcal{X}$'s dependency parse, 
and then regard the combination of the context features and their in-bound dependency types as the corresponding dependency information.
For example, as illustrated in Figure \ref{fig:knowledge}(c), for ``\textit{Salt}'', its context features are ``\textit{Salt}'' and ``\textit{City}'' (the governor of ``\textit{Salt}''), and their corresponding dependency information are ``\textit{Salt\_compound}'' and ``\textit{City\_root}''.\footnote{
Note that, in this case, we do not have context features selected from the dependents since ``\textit{Salt}'' do not have any dependents according to the dependency parse result.}

%

\vspace{0.2cm}

As a result, for each type of syntactic inforamtion, we obtain a list of context features and a list of syntactic information instances, which are modeled by a KVMN module to enhance input text representation and thus improve model performance.


\subsection{KVMN for Syntactic Information}


Since the syntactic information is obtained from off-the-shelf toolkits, it is possible that there is noise in the extracted syntactic information, which may hurt model performance if it is not leveraged appropriately.
Inspired by the studies that use KVMN and its variants to weight and leverage extra features to enhance model performance in many NLP tasks \cite{miller2016key,mino-etal-2017-key,xu-etal-2019-enhancing,tian2020improving}, for
%
%
%
each type of the syntactic information (denoted as $c$),
we propose a KVMN module ($\mathcal{M}^{c}$) to model the pair-wisely organized context features and the syntactic information instances.
Specifically, for each $x_{i}$ in the input, we firstly map its context features and the syntactic information to keys and values in the KVMN, which are denoted by $\mathcal{K}^{c}_{i} = [k^{c}_{i,1}, \ldots, k^{c}_{i,j}, \ldots, k^{c}_{i, m_i}]$ and $\mathcal{V}^{c}_i = [v^{c}_{i,1}, \ldots, v^{c}_{i,j}, \ldots, v^{c}_{i, m_i}]$, respectively, with $m_i$ the number of context features for $x_i$.
Next, we use two matrices to map them to their embeddings, with $\mathbf{e}^{k_{c}}_{i,j}$ referring to the embedding of $k^{c}_{i,j}$ and $\mathbf{e}_{i,j}^{v_c}$ for $v^{c}_{i,j}$, respectively.
Then, for each token $x_{i}$ and its associated context features $\mathcal{K}^{c}_{i}$ and syntactic information $\mathcal{V}^{c}_{i}$,
the weight assigned to the syntactic information $v^{c}_{i,j}$ is computed by
\begin{equation} \label{eq: p_ij}
\setlength\abovedisplayskip{6pt}
\setlength\belowdisplayskip{6pt}
    p^{c}_{i,j} = \frac{exp(\mathbf{h}_i\cdot\mathbf{e}_{i,j}^{k_c})}{\sum_{j=1}^{m_i}exp(\mathbf{h}_i\cdot\mathbf{e}_{i,j}^{k_c})}
\end{equation}
where $\mathbf{h}_{i}$ is the hidden vector for $x_{i}$ obtained from the context encoder.
%
%
Afterwards, we apply the weights $p^{c}_{i,j}$ to their corresponding syntactic information $v^{c}_{i,j}$ by
\begin{equation}
\setlength\abovedisplayskip{6pt}
\setlength\belowdisplayskip{6pt}
    \mathbf{s}^{c}_{i} = \sum_{j=1}^{m_i} p^{c}_{i,j} \mathbf{e}_{i,j}^{v_{c}}
\end{equation}
where $\mathbf{s}^{c}_{i}$ is the output of $\mathcal{M}^{c}$, containing the weighted syntactic information in type $c$.
Therefore, KVMN ensures that the syntactic information are weighted according to their corresponding context features, so that important information could be distinguished and leveraged accordingly.


\begin{table*}[t]
    \centering
    \small
    \begin{sc}
    \begin{tabular}{L{0.7cm}R{0.8cm}R{0.8cm}R{0.8cm}R{0.8cm}R{0.8cm}R{0.8cm}R{0.8cm}R{0.8cm}R{0.8cm}R{0.8cm}R{0.8cm}R{0.8cm}
    }
    \toprule
    \multirow{4}{*}{\textbf{Type}}&  
    \multicolumn{6}{c}{\textbf{English}}&\multicolumn{6}{c}{\textbf{Chinese}}\cr
    \cmidrule(lr){2-7}\cmidrule(lr){8-13}
    &\multicolumn{2}{c}{\textbf{ON5e}}&\multicolumn{2}{c}{\textbf{WN16}}&\multicolumn{2}{c}{\textbf{WN17}}&\multicolumn{2}{c}{\textbf{ON4c}}&\multicolumn{2}{c}{\textbf{RE}}&\multicolumn{2}{c}{\textbf{WE}}\cr
    \cmidrule(lr){2-3} \cmidrule(lr){4-5} \cmidrule(lr){6-7} \cmidrule(lr){8-9} \cmidrule(lr){10-11} \cmidrule(lr){12-13} 
    & \multicolumn{2}{c}{$\#$ T. = 18} & \multicolumn{2}{c}{$\#$ T. = 10} & \multicolumn{2}{c}{$\#$ T. = 6} & \multicolumn{2}{c}{$\#$ T. = 4} & \multicolumn{2}{c}{$\#$ T. = 8} & \multicolumn{2}{c}{$\#$ T. = 4} \cr
    \cmidrule(lr){2-3} \cmidrule(lr){4-5} \cmidrule(lr){6-7} \cmidrule(lr){8-9} \cmidrule(lr){10-11} \cmidrule(lr){12-13} 
        &$\#$ S.    &$\#$ E.    &$\#$ S.    &$\#$ E.    &$\#$ S.    &$\#$ E.    &$\#$ S.    &$\#$ E.    &$\#$ S.    &$\#$ E.    &$\#$ S.    &$\#$ E.    \cr  
    \midrule
    Train & 59.9K & 81.8K & 2.4K & 1.5K & 3.4K & 2.0K & 15.7K & 13.4K & 3.8K & 13.4K & 1.4K & 1.9K \\
    Dev   & 8.5K & 11.1K & 1.0K & 0.7K & 1.0K & 0.8K & 4.3K & 7.0K & 0.5K & 1.5K & 0.3K & 0.4K \\
    Test  & 8.3K & 11.3K & 3.9K & 3.5K & 1.3K & 1.1K & 4.3K & 7.7K & 0.5K & 1.6K & 0.3K & 0.4K \\
    
    \bottomrule
    \end{tabular}
    \end{sc}
    \vspace{-0.2cm}
    \caption{
    Statistics of all datasets with respect to the number of NE types (T.), sentences (S.), and total NEs (E.).}
    \label{tab:dataset}
    \vskip -0.5em
\end{table*}


\subsection{The Syntax Attention}

Upon encoding each type of syntactic information by KVMN, one can assemble different types of them with an overall representation.
The most straightforward way of doing so is to concatenate the encoding from each type by
\begin{equation}
    \setlength\abovedisplayskip{6pt}
    \setlength\belowdisplayskip{6pt}
    \mathbf{s}_i = \mathop{\oplus} \limits_{c \in \mathcal{C}} \mathbf{s}_i^c
\end{equation}
%
%
where
$\mathbf{s}_i$ is the aggregated results of $\mathbf{s}_i^c$, the embedding for each syntactic type from $\mathcal{M}^{c}$.
However, given the fact that different syntactic information may conflict to each other, it is expected to have a more effective way to combine them.

%
Motivated by studies that selectively leverage different features by assigning different weights to them \cite{kumar-etal-2018-knowledge,higashiyama-etal-2019-incorporating,tian-etal-2020-joint,tian-etal-2020-suppertagging},
we propose a syntax attention for the syntactic information ensemble.
Particularly,
for each syntactic type $c$, we firstly concatenate $\mathbf{s}^{c}_{i}$ with $\mathbf{h}_{i}$ and use the resulting vector to compute the weight $q^{c}_{i}$ for $\mathbf{s}^{c}_{i}$:
\begin{equation}
\setlength\abovedisplayskip{6pt}
\setlength\belowdisplayskip{6pt}
    q_{i}^{c} = \sigma(\mathbf{W}_{q}^{c}\cdot(\mathbf{h}_i\oplus \mathbf{s}_i^{c}) + b_q^c)
\end{equation}
where $\mathbf{W}_{q}^{c}$ and $b^{c}_{q}$ are trainable vector and variable, respectively, and $\sigma$ is the \textit{sigmoid} function.
%
Then, a \textit{softmax} function is applied over all types of syntactic information to compute their corresponding attentions $a^{c}_{i}$, which is formalized by
\begin{equation} \label{eq: a_i}
\setlength\abovedisplayskip{6pt}
\setlength\belowdisplayskip{6pt}
    a_i^c = \frac{exp(q_i^c)}{\sum_{c \in \mathcal{C}} exp(q_i^c)}
\end{equation}
In the last, we apply the weights to their corresponding encoded syntactic information vectors by
\begin{equation}
\setlength\abovedisplayskip{6pt}
\setlength\belowdisplayskip{6pt}
    \mathbf{s}_i = \sum_{c \in \mathcal{C}} a_i^c \mathbf{s}_i^c 
\end{equation}
where $\mathbf{s}_{i}$ is the output of the syntax attention of different syntactic information types.

\subsection{The Gate Mechanism}

To enhance NER with
the syntactic information encoded by KVMN and combined by syntax attention,
we propose a gate mechanism ($\mathcal{GM}$) to incorporate it to the backbone NER model,
where we expect such mechanism could dynamically weight and decide how to leverage the syntactic information in labeling NEs.
In detail,
we propose 
a \textit{reset} function $\mathbf{r}_{i}$ to evaluate the encodings from the context encoder and the syntax attention by
%
%
%
\begin{equation} \label{eq: r}
\setlength\abovedisplayskip{6pt}
\setlength\belowdisplayskip{6pt}
    \mathbf{r}_{i} = \sigma(\mathbf{W}_{r_1} \cdot \mathbf{h_i} + \mathbf{W}_{r_2}\cdot \mathbf{s_i} + \mathbf{b}_r)
\end{equation}
where $\mathbf{W}_{r_1}$, $\mathbf{W}_{r_2}$ are trainable matrices and $\mathbf{b}_r$ the bias term,
and use
\begin{equation}
\setlength\abovedisplayskip{6pt}
\setlength\belowdisplayskip{6pt}
     \mathbf{o}_i = [\mathbf{r}_{i} \circ \mathbf{h}_i] \oplus [(\textbf{1} - \mathbf{r}_{i}) \circ \mathbf{s}_i]
\end{equation}
to control the contribution of them,
where $\mathbf{o}_{i}$ is the output of the gate mechanism corresponding to input $x_i$,
$\mathbf{1}$ is a 1-vector with its dimension matching $\mathbf{h}_{i}$ and $\circ$ the element-wise multiplication operation.

\subsection{Encoding and Decoding for NER}


To provide $\mathbf{h}_{i}$ to KVMN,
we adopt Adapted-Transformer\footnote{The Adapted-Transformer additionally models direction and distance information of the input, which are demonstrated to be useful for NER comparing to the vanilla Transformer.} \cite{DBLP:journals/corr/abs-1911-04474} as the context encoder in this work.
So that the encoding of the input text can be formalized as
%
%
%
\begin{equation}
\setlength\abovedisplayskip{6pt}
\setlength\belowdisplayskip{6pt}
    \mathbf{H} = \text{Adapted-Transformer}(\mathbf{E})
\end{equation}
where $\mathbf{H} = [\mathbf{h}_{1}, \mathbf{h}_{2}, \cdots, \mathbf{h}_{i}, \cdots, \mathbf{h}_{n}]$ and $\mathbf{E} = [\mathbf{e}_{1}, \mathbf{e}_{2}, \cdots, \mathbf{e}_{i}, \cdots, \mathbf{e}_{n}]$ are lists of hidden vectors and embeddings of $\mathcal{X}$, respectively.
Note that, since pre-trained embeddings contain context information learned from large-scale corpora, and
different types of them 
may carry heterogeneous context information learned from different algorithms and corpora, 
%
we incorporate multiple pre-trained embeddings by direct concatenating them in the input:
%
%
\begin{equation}
\setlength\abovedisplayskip{6pt}
\setlength\belowdisplayskip{6pt}
    \mathbf{e}_{i} = \mathop{\oplus} \limits_{z \in \mathcal{Z}} \mathbf{e}_{i}^{z}
\end{equation}
where $\mathbf{e}_{i}$ is the final word representation to feed the context encoder; 
$\mathbf{e}_{i}^{z}$ represents the word embedding of $x_{i}$ in embedding type $z$ and $\mathcal{Z}$ the set of all embedding types.

For the output, upon the receiving of $\mathbf{o}_{i}$,
%
a trainable matrix $\mathbf{W}_{o}$ is used to align its dimension to the output space by $\mathbf{u}_{i} = \mathbf{W}_{o} \cdot \mathbf{o}_i$.
Finally, we apply a conditional random field (CRF) decoder to predict the labels $\hat{y}_{i} \in \mathcal{T}$ (where $\mathcal{T}$ is the set with all NE labels) in the output sequence $\mathcal{\hat{Y}}$ by
\begin{equation} \label{eq:crf}
\setlength\abovedisplayskip{6pt}
\setlength\belowdisplayskip{6pt}
    \hat{y}_{i} = \underset{y_{i} \in \mathcal{T}}{\arg \max} \frac{ exp(\mathbf{W}_{c} \cdot \mathbf{u}_{i} + \mathbf{b}_{c})}
                    {\sum_{y_{i-1}y_{i}} exp(\mathbf{W}_{c} \cdot \mathbf{u}_{i} + \mathbf{b}_{c})}
\end{equation}
where $\mathbf{W}_{c}$ and $\mathbf{b}_{c}$ are trainable parameters to model the transition for $y_{i-1}$ to $y_{i}$.

\section{Experimental Settings}
\label{exp}


\subsection{Datasets}



%

In our experiments, we use three English benchmark datasets, i.e., OntoNotes 5.0 (ON5e) \cite{DBLP:conf/conll/PradhanMXNBUZZ13} , WNUT-16 (WN16), WNUT-17 (WN17) \cite{DBLP:conf/aclnut/DerczynskiNEL17}, and three Chinese datasets, i.e., OntoNotes 4.0 (ON4c) \cite{weischedel2011ontonotes}, Resume (RE) \cite{DBLP:conf/acl/ZhangY18}, Weibo (WE) \cite{DBLP:conf/emnlp/PengD15}.\footnote{Among these datasets, ON5e and ON4c are multi-lingual datasets. We follow \newcite{DBLP:journals/corr/abs-1911-04474} by extracting the corresponding English and Chinese part from them.}
These datasets come from a wide range of sources
so that we are able to comprehensively evaluate our approach with them.
In detail,
WN16 and WN17 are constructed from social media;
ON5e consists of mixed sources, such as telephone conversation, newswire, etc.;
ON4c is from news domain;
RE and WE are extracted from Chinese online resources.
For all datasets, we use their original splits and 
the statistics of 
them
with respect to the number of entity types (\textsc{\# T.}), sentences (\textsc{\# S.}) and total entities (\textsc{\# E.}) in the train/dev/test sets are reported in Table \ref{tab:dataset}.

\begin{table*}[t]
    \centering
    \begin{sc}
    \begin{tabular}{C{1.2cm}C{1.2cm}C{1.2cm}C{1.4cm}C{1.4cm}C{1.4cm}C{1.35cm}C{1.35cm}C{1.35cm}}
        \toprule
        \multicolumn{3}{c}{\textbf{Syntactic \ \  Information}}&\multirow{2}{*}{\textbf{ON5e}}&\multirow{2}{*}{\textbf{WN16}}&\multirow{2}{*}{\textbf{WN17}}&\multirow{2}{*}{\textbf{ON4c}}&\multirow{2}{*}{\textbf{RE}}&\multirow{2}{*}{\textbf{WE}}\cr
        \cmidrule{1-3}
        \textbf{POS.}&\textbf{Con.}&\textbf{Dep.}\cr
        \midrule
            &&                                  & 89.32 & 53.81 & 48.96 & 79.04 & 95.84 & 67.79 \cr
            $\surd$&&                           & 89.51 & 53.94 & 49.68 & 79.53 & 96.09 & \textbf{68.76} \cr
            &$\surd$&                           & \textbf{89.64} & \textbf{54.59} & \textbf{49.82} & 79.76 & \textbf{96.11} & 68.11 \cr
            &&$\surd$                           & 89.58 & 54.37 & 49.47 & \textbf{80.03} & 96.02 & 68.64 \cr
        \bottomrule
    \end{tabular}
    \end{sc}
    \vspace{-0.2cm}
    \caption{
    $F1$ scores of the baseline model and ours enhanced with
    different types of syntactic information (``\begin{sc}POS.\end{sc}'', ``\begin{sc}Con.\end{sc}'' and ``\begin{sc}Dep.\end{sc}'' refer to POS labels, syntactic constituents and dependency relations, respectively).
    }
    \label{tab:kv}
    \vskip -0.1em
\end{table*}

\begin{table*}[t]
    \centering
    \begin{sc}
    \begin{tabular}{L{1.0cm}C{1.2cm}C{1.2cm}C{1.2cm}C{1.15cm}C{1.15cm}C{1.15cm}C{1.15cm}C{1.1cm}C{1.1cm}}
        \toprule
        \multirow{2}{*}{\textbf{Type}} &\multicolumn{3}{c}{\textbf{Syntactic \ \ Information}}&\multirow{2}{*}{\textbf{ON5e}}&\multirow{2}{*}{\textbf{WN16}}&\multirow{2}{*}{\textbf{WN17}}&\multirow{2}{*}{\textbf{ON4c}}&\multirow{2}{*}{\textbf{RE}}&\multirow{2}{*}{\textbf{WE}}\cr
        \cmidrule{2-4}
        &\textbf{POS.}&\textbf{Con.}&\textbf{Dep.}\cr
           \midrule

               \multirow{4}{*}{$\mathcal{DC}$}&$\surd$&$\surd$&         & 89.61 & 54.11 & 49.61 & 79.61 & 95.72 & 68.27  \cr
               &$\surd$&&$\surd$                                        & 89.56 & 54.03 & 49.74 & 79.83 & 96.11 & 68.51 \cr
               &&$\surd$&$\surd$                                        & 89.60 & 54.26 & 49.58 & 79.89 & 96.08 & 68.36 \cr
               &$\surd$&$\surd$&$\surd$                                 & 89.62 & 54.41 & 49.63 & 79.81 & 95.31 & 68.49 \cr
                 \midrule
         \multirow{4}{*}{$\mathcal{SA}$}&$\surd$&$\surd$&               & 89.68 & 54.68 & 49.81 & 79.92 & 96.19 & 68.94 \cr
                        &$\surd$&&$\surd$                               & 89.76 & 54.61 & 49.89 & 80.29 & 96.23 & 69.01 \cr
                        &&$\surd$&$\surd$                               & 89.78 & 54.56 & 49.96 & 80.41 & 96.31 & 68.76  \cr
                        &$\surd$&$\surd$&$\surd$                        & \textbf{89.86} & \textbf{54.79} & \textbf{50.21} & \textbf{80.65} & \textbf{96.43} & \textbf{69.37} \cr
        \bottomrule
    \end{tabular}
    \end{sc}
    \vspace{-0.2cm}
    \caption{
    $F1$ scores of our models with 
    different combinations of syntactic information.
    ``\begin{sc}Type\end{sc}'' indicates how they are combined, where ``$\mathcal{DC}$'' and ``$\mathcal{SA}$'' refer to direct concatenation and syntax attention, respectively.}
    \label{tab:syn-attn}
    \vskip -0.5em
\end{table*}

\subsection{Implementation}

To label NEs, we use the \textbf{BIOES} tagging scheme instead of the standard \textbf{BIO} scheme for the reason that previous studies have shown optimistic improvement with this scheme \cite{DBLP:conf/naacl/LampleBSKD16,DBLP:journals/corr/abs-1911-04474}.
%
For the text input, 
we use three types of embeddings for each language by default.
Specifically, for English, we use Glove (100-dimension)\footnote{We download the Glove.6B embedding from \url{https://nlp.stanford.edu/projects/glove/}} \cite{DBLP:conf/emnlp/PenningtonSM14}, ELMo \cite{DBLP:conf/naacl/PetersNIGCLZ18}, and the BERT-cased large\footnote{We obtain the pre-trained BERT from \url{https://github.com/google-research/bert}.} \cite{DBLP:conf/naacl/DevlinCLT19} (the derived embeddings for each word);
for Chinese, we use pre-trained character and bi-gram embeddings\footnote{We obtain the embeddings from \url{https://github.com/jiesutd/LatticeLSTM}.} released by \citet{DBLP:conf/acl/ZhangY18} (denoted as Giga), Tencent Embedding\footnote{We use the official release from \url{https://ai.tencent.com/ailab/nlp/embedding.html}.} \cite{DBLP:conf/naacl/SongSLZ18}, and ZEN\footnote{We use the pre-trained ZEN-base downloaded from \url{https://github.com/sinovation/ZEN}. Note that we do not use the Chinese BERT since ZEN performs better across three Chinese datasets. For reference, we report the results of using BERT in Appendix A.} \cite{DBLP:journals/corr/zen}.
%
For both BERT and ZEN, we follow their default settings, i.e., 24 layers of self-attention with 1024 dimensional embeddings for BERT-large and 12 layers of self-attention with 768 dimensional embeddings for ZEN-base.
%
For syntactic information, we use the Stanford CoreNLP Toolkit\footnote{We use its 3.9.2 version downloaded from \url{https://stanfordnlp.github.io/CoreNLP/}.} \cite{DBLP:conf/acl/ManningSBFBM14} to produce the aforementioned three types of syntactic information, i.e. POS labels, syntactic constituents, and dependency relations, for each input text.
In the context encoding layer, we use a two-layer Adapted-Transformer encoder\footnote{We also try other encoders (i.e., Bi-LSTM and Transformer) and report their results in Appendix B for reference.} with 128 hidden units and 12 heads and set the dropout rate to 0.2.
For the memory module, all key and value embeddings are initialized randomly.
%
During the training process, we fix all pre-trained embeddings and use Adam \cite{DBLP:journals/corr/KingmaB14} to optimize negative log-likelihood loss function with the learning rate set to $\eta = 0.0001$, $\beta_1 = 0.9$ and $\beta_2 = 0.99$. 
In all experiments, we run a maximum of 100 epochs with the batch size of 32 and tune the hyper-parameters on the development set.\footnote{We report the hyper-parameter settings of different models as well as the best one in Appendix C.}
%
The model that achieves the highest performance on the development set is evaluated on the test set with respect to the $F1$ scores obtained from the official \textit{conlleval} toolkits\footnote{\url{https://www.clips.uantwerpen.be/conll2000/chunking/conlleval.txt}.}.

\begin{table*}[t]
    \centering
    \small
    \begin{sc}
    \begin{tabular}{L{1.0cm}C{1.22cm}C{1.22cm}C{1.22cm}C{1.18cm}C{1.18cm}C{1.18cm}C{1.18cm}C{1.18cm}C{1.18cm}}
        \toprule
        \multirow{2}{*}{\textbf{$\mathcal{GM}$}}&\multicolumn{3}{c}{\textbf{Syntactic \ \ Information}}&\multirow{2}{*}{\textbf{ON5e}}&\multirow{2}{*}{\textbf{WN16}}&\multirow{2}{*}{\textbf{WN17}}&\multirow{2}{*}{\textbf{ON4c}}&\multirow{2}{*}{\textbf{RE}}&\multirow{2}{*}{\textbf{WE}}\cr
        \cmidrule{2-4}
        &\textbf{POS.}&\textbf{Con.}&\textbf{Dep.}\cr
        \midrule
        & $\surd$ & $\surd$ &                   & 89.68 & 54.68 & 49.81 & 79.92 & 96.19 & 68.94 \cr
        $\surd$ & $\surd$ & $\surd$ &           & 90.09 & 54.92 & 50.28 & 80.31 & 96.51 & 69.31 \cr
        \cmidrule(lr){1-10}
         & $\surd$ & & $\surd$                  & 89.76 & 54.61 & 49.89 & 80.29 & 96.23 & 69.01 \cr
        $\surd$ & $\surd$ & & $\surd$           & 90.08 & 54.78 & 50.16 & 80.64 & 96.47 & 69.47 \cr
        \cmidrule(lr){1-10}
         & & $\surd$ & $\surd$                  & 89.78 & 54.56 & 49.96 & 80.41 & 96.31 & 68.76 \cr
        $\surd$ & & $\surd$ & $\surd$           & 90.11 & 54.96 & 50.36 & 80.87 & 96.51 & 69.24 \cr
        \cmidrule(lr){1-10}
         & $\surd$ & $\surd$ & $\surd$          & 89.86 & 54.79 & 50.21 & 80.65 & 96.43 & 69.37 \cr
        $\surd$ & $\surd$ & $\surd$ & $\surd$   & \textbf{90.32}& \textbf{55.14} & \textbf{50.68} & \textbf{81.18} & \textbf{96.62} & \textbf{69.78} \cr
        \bottomrule
    \end{tabular}
    \end{sc}
    \vspace{-0.2cm}
    \caption{
    $F1$ scores of our models
    with and without applying the gate mechanism (``$\mathcal{GM}$'') when different syntactic information are combined by syntactic attention.
    }
    \label{tab:effect-gate}
\end{table*}


\begin{table*}[t]
    \centering
    \begin{small}
    \begin{sc}
    \begin{tabular}{L{3.7cm}C{0.9cm}C{0.9cm}C{0.9cm}|L{3.6cm}C{0.9cm}C{0.9cm}C{0.8cm}}
    \toprule
    \multicolumn{4}{c}{\textbf{English}} & \multicolumn{4}{c}{\textbf{Chinese}} \\
    \cmidrule(lr){1-4}\cmidrule(lr){5-8}
    \textbf{Model} & \textbf{ON5e} & \textbf{WN16} & \textbf{WN17} & \textbf{Model} & \textbf{ON4c}  & \textbf{RE} & \textbf{WE} \\
    \midrule
    \newcite{DBLP:journals/tacl/ChiuN16}    & 86.12 & - & - & \newcite{DBLP:conf/acl/ZhangY18}              & 73.88 & - & 58.79 \\
    $^\dag$\newcite{DBLP:journals/bioinformatics/LuoYYZWLW18}  & 88.79 & 51.26 & 48.63 & \newcite{DBLP:journals/corr/abs-1911-04474}& 72.43 & 95.00 & 58.17 \\
    $^\dag$\newcite{DBLP:journals/bioinformatics/DangLNV18}    & 88.91 & 51.84 & 48.12 & \newcite{DBLP:conf/emnlp/GuiZZPFWH19} & 74.89 & 95.37 & 60.21 \\
    \newcite{DBLP:conf/coling/AkbikBV18}     & 89.30 & - & - & \newcite{DBLP:conf/naacl/ZhuW19} & 73.64 & 94.94 & 59.31 \\
    \newcite{DBLP:conf/emnlp/JieL19}        & 89.88 & - & - & \newcite{DBLP:conf/ijcai/GuiM0ZJH19} & 74.45 & 95.11 & 59.92 \\
    \newcite{DBLP:journals/corr/abs-1911-04474}      & 89.78 & 54.06 & 48.98 & \newcite{DBLP:conf/naacl/LiuXXSZ19} & 74.43 & 95.21 & 59.84 \\
    $^*$\newcite{DBLP:conf/naacl/DevlinCLT19}        & 89.16 & 54.36 & 49.52 & \newcite{DBLP:conf/emnlp/SuiCLZL19} & 74.79 & - & 63.09 \\
    \newcite{DBLP:conf/acl/ZhouZJZFGK19}    & - & 53.43 & 42.83 & \newcite{DBLP:conf/acl/DingXZLLS19}     & 76.00 & - & 59.50 \\
    \newcite{DBLP:conf/naacl/AkbikBV19}     & - & - & 49.59 & $^*$\newcite{DBLP:conf/nips/MengWWLNYLHSL19}    & 80.62 & 96.54 & 67.60 \\
    \newcite{DBLP:conf/ijcai/DaiF019}       & 89.83 & - &  - & \newcite{DBLP:conf/cikm/XuWHL19}         & - & - & 68.93 \\
    \newcite{DBLP:conf/acl/LiuYL19}         & 89.94 & - & - & \newcite{DBLP:conf/acl/simplify} & 75.54 & 95.59 & 61.24 \\
    $^*$\newcite{DBLP:conf/aaai/LuoXZ20} & 90.30 & - & - & $^*$\newcite{DBLP:conf/seke/SLK} & 80.20 & 95.80 & 64.00 \\
    \midrule
    \textbf{Ours}   & \textbf{90.32}& \textbf{55.14}& \textbf{50.68} & \textbf{Ours} & \textbf{81.18}& \textbf{96.62}& \textbf{69.78} \\
    \bottomrule
    \end{tabular}
    \end{sc}
    \end{small}
    \vspace{-0.2cm}
    \caption{
    Comparison of $F1$ scores of our best performing model (i.e. the full model with attentive ensemble of all syntactic information)
    with that reported in previous studies on all English and Chinese benchmark datasets.
    ``*'' indicates the studies using BERT as the text encoder; ``$\dag$'' means the results are our runs of their models.
    }
    \label{tab:comparison}
    \vskip -1.0em
\end{table*}

\section{Experimental Results}
\label{res}




\subsection{Effect of Key-Value Memory Networks}

To explore how different syntactic information helps NER, we run the baselines without syntactic information and the ones with each type of syntactic information through KVMN.\footnote{
Syntax attention and the gate mechanism are not applied.}
Experimental results ($F1$) are reported in Table \ref{tab:kv} for all datasets. 
%

It is observed from the results that the models with syntactic information outperform the baseline in all cases,
which demonstrates the effectiveness of using KVMN in our model.
In addition, it is also noticed that the best performed model is not exactly the same one across different datasets, which indicates the contributions of different syntactic information vary in those datasets.
For example, in most datasets, models using syntactic constituents achieve the best results, which can be explained by that syntactic constituents provide important cues of NE chunks.
As a comparison,
POS labels are the most effective syntactic information for WE dataset, which can be attributed to the natural of the dataset that most sentences in social media are not formally written, so that their parsing results could be inaccurate and mislead the NER process.

\subsection{Effect of Syntax Attention}


To examine the effectiveness of syntax attention ($\mathcal{SA}$), we compare 
it with another strategy through
direct concatenation ($\mathcal{DC}$) of the KVMN output to model ouptut.
%
The results are reported in
Table \ref{tab:syn-attn} with applying all combinations of different syntactic information by $\mathcal{DC}$ and $\mathcal{SA}$.
%

There are several observations.
First,
interestingly,
compared to the results in Table \ref{tab:kv},
%
direct concatenation of different syntactic information hurts NER performance in most cases.
For example, on the RE dataset, the ensemble of all types of syntactic information 
through $\mathcal{DC}$ obtains the worst results compared to all other results with integrating less information under the same setting.
The reason behind this phenomenon could be that different syntactic information may provide conflict cues to NE tags and thus
result in inferior performance.
Second,
on the contrary,
$\mathcal{SA}$ is able to improve NER with integrating multiple types of syntactic information,
where consistent improvements are observed among all datasets when more types of syntactic information are incorporated.
As a result, the best results are achieved by the model using all types of syntactic information. 
This observation suggests that 
the syntax attention is able to weight different syntactic information and distinguish important ones from others,
thus alleviate possible conflicts of them when labeling entities.

\begin{table*}[t]
    \centering
    \small
    \begin{sc}
    \begin{tabular}{C{0.8cm}C{0.8cm}C{0.8cm}C{0.9cm}C{0.9cm}C{0.9cm}|C{0.8cm}C{1.2cm}C{0.8cm}C{1.0cm}C{1.0cm}C{1.0cm}}
        \toprule
        \multicolumn{6}{c}{\textbf{English}} & \multicolumn{6}{c}{\textbf{Chinese}} \cr
        \cmidrule(lr){1-6}\cmidrule(lr){7-12}
        
        \multicolumn{3}{c}{\textbf{Embeddings}} & \multirow{2}{*}{\textbf{ON5e}} & \multirow{2}{*}{\textbf{WN16}} & \multirow{2}{*}{\textbf{WN17}} & \multicolumn{3}{c}{\textbf{Embeddings}} & \multirow{2}{*}{\textbf{ON4c}} & \multirow{2}{*}{\textbf{RE}} & \multirow{2}{*}{\textbf{WE}} \cr
        \cmidrule(lr){1-3} \cmidrule(lr){7-9}
        
        \textbf{Glove} & \textbf{ELMo} & \textbf{BERT} &&& & \textbf{Giga} & \textbf{Tencent} & \textbf{ZEN} &&& \cr
        \midrule
        $\surd$ & &                 & 89.37 & 47.92 & 43.24 & $\surd$ & &             & 72.11 & 94.99 & 61.94 \\      
        & $\surd$ &                 & 89.71 & 53.96 & 47.92 & & $\surd$ &             & 73.54 & 95.21 & 63.06 \\      
        & & $\surd$                 & 89.53 & 53.74 & 48.74 & & & $\surd$             & 80.06 & 95.98 & 68.84 \\
        $\surd$ & $\surd$ &         & 89.91 & 54.36 & 48.21 & $\surd$ & $\surd$ &     & 74.86 & 95.46 & 63.96 \\
        $\surd$ & & $\surd$         & 89.82 & 54.16 & 49.61 & $\surd$ & & $\surd$     & 80.49 & 96.24 & 68.94 \\
        & $\surd$ & $\surd$         & 90.13 & 54.92 & 50.12 & & $\surd$ & $\surd$     & 80.81 & 96.41 & 69.42 \\
        $\surd$ & $\surd$ & $\surd$ & \textbf{90.32} & \textbf{55.14} & \textbf{50.68} & $\surd$ & $\surd$ & $\surd$ & \textbf{81.18} & \textbf{96.62} & \textbf{69.78} \\
        \bottomrule
    \end{tabular}
    \end{sc}
    \vspace{-0.2cm}
    \caption{
    Experimental results ($F1$ scores) of our best performing model (i.e., the full model with attentive ensemble of all syntactic information) using different pre-trained embeddings and their combinations as input.
    }
    \label{tab:embedding}
\end{table*}

\subsection{Effect of the Gate Mechanism}


We experiment our model under its best setting (i.e., $\mathcal{SA}$ over all combinations of syntactic information with KVMN) with and without the gate mechanism to investigate its effectiveness of actively controlling the information flow from the context encoder and $\mathcal{SA}$.
%
The results 
are presented in
Table \ref{tab:effect-gate}, where the ones without using the gate mechanism are obtained directly from Table \ref{tab:syn-attn}.\footnote{The results of those models on the development sets of all datasets are reported in \textcolor{blue}{Appendix D.}}
It is clearly shown that in all cases, the model with gate mechanism 
achieves superior performance to the other one without it.
These results suggest that the importance of the information from the context encoder and $\mathcal{SA}$ varies, so that the proposed gate mechanism is effective in balancing them.

\subsection{Comparison with Previous Studies}
\label{sec: comp}

To further illustrate the effectiveness of our models,
we compare the best performing one, i.e., the last line in Table \ref{tab:effect-gate}, with the results from previous studies.
The results are shown in Table \ref{tab:comparison}, where 
our approach outperforms previous models with BERT encoder (marked by ``*'')
and achieves state-of-the-art results on all English and Chinese datasets.
This observation indicates that incorporate different embeddings as input is more effective than directly using pre-trained models.
%
%
In addition,
compared to some previous studies \cite{DBLP:journals/bioinformatics/LuoYYZWLW18,DBLP:journals/bioinformatics/DangLNV18}\footnote{\citet{DBLP:journals/bioinformatics/LuoYYZWLW18} and \citet{DBLP:journals/bioinformatics/DangLNV18} do not report their results on all general domain benchmark datasets, because the focus of their studies is biomedical NER.
Therefore, we report our runs of their method in Table \ref{tab:comparison} (marked by ``$\dag$'').}
that leverage multiple types of syntactic information
by regarding the information as gold references and directly concatenating their embeddings with word embeddings,
our approach has its superiority by using attentive ensemble through KVMN, syntax attention, and the gate mechanism to selectively learn from different syntactic information according to their contribution to NER, where such multi-phase strategy of attentive ensemble guarantees the appropriateness of learning them in a reasonable manner.
%

\begin{figure*}[t]
    \centering
    \small
    \includegraphics[width=\textwidth, trim=0 20 0 0]{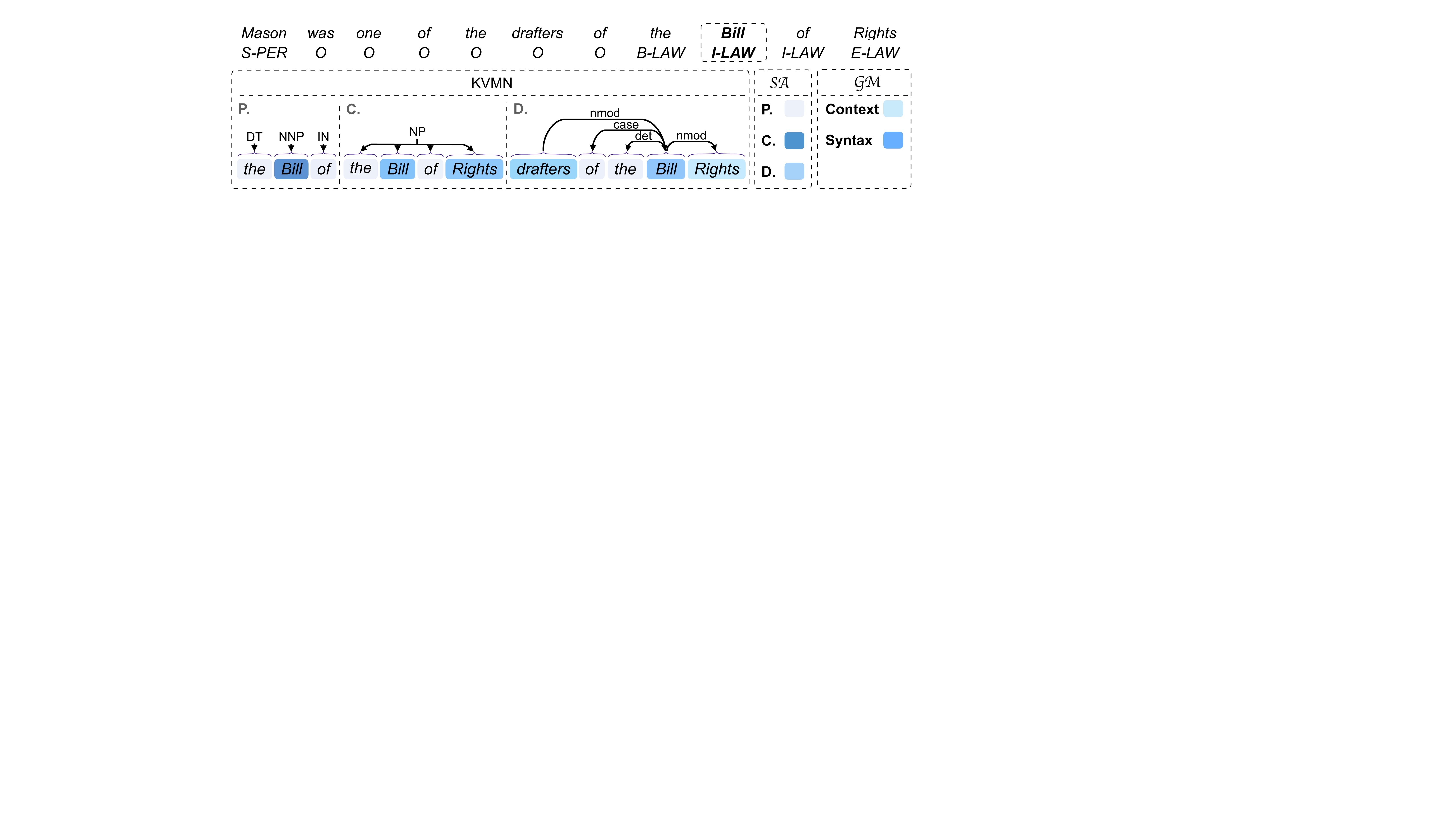}
    \caption{
    An illustration of how our model encodes syntactic information through KVMN, weights them by syntax attention ($\mathcal{SA}$) and learns from the gate mechanism ($\mathcal{GM}$),
    where the weights for different features and information types are visualized.
    The example sentence is shown at the top with the gold NE tags for each word marked below.
    The weights assigned to different syntactic information for ``\textit{Bill}'' in KVMN, $\mathcal{SA}$, and $\mathcal{GM}$ are highlighted with colors,
    where the darker colors referring to higher values.
    }
    \label{fig:case}
    \vskip -0.8em
\end{figure*}


\section{Analyses}

\subsection{Effect of Different Word Embeddings}


Neural models are sensitive to input embeddings, which is also true for our approach.
Consider that different types of embeddings carry contextual information learned from various corpora and algorithms, we explore the effect of those embeddings when they are used separately or combined as the input.
The experiment is performed on our best model (i.e., KVMN+$\mathcal{SA}$+$\mathcal{GM}$ on all syntactic information),
%
with the results reported in Table \ref{tab:embedding}.
%
It is clearly observed
that for all English and Chinese datasets, the model with all three embeddings achieves the best performance and its performance drops consistently when more types of embeddings are excluded.
It is confirmed that different types of embedding do provide complement context information to enhance the understanding of input texts for NER.
Particularly, although contextualized embeddings (i.e., ELMo, BERT, and ZEN) show significantly better performance than others (especially on Chinese),
combining them with static embeddings still provide further improvement on the $F1$ score of NER systems.

\subsection{Case Study}

%
To better understand how attentive ensemble of syntactic information
helps NER, we conduct a case study for the word ``\textit{Bill}'' in an example sentence ``\textit{Mason was one of the drafters of the Bill of Rights}'' from the ON5e dataset.
%
Figure \ref{fig:case} visualizes the weights for different context features in KVMN,
as well as the weights from
$\mathcal{SA}$
and $\mathcal{GM}$,
where darker colors refer to higher weights.\footnote{For the weights, we visualize $p^{c}_{i,j}$ in Eq. (\ref{eq: p_ij}) for KVMN, $a_{i}^{c}$ in Eq. (\ref{eq: a_i}) for $\mathcal{SA}$, and $||\mathbf{r}_{i}||_{2}$ for $\mathbf{r_i}$ in Eq. (\ref{eq: r}) for $\mathcal{GM}$.}

Interestingly,
in this case,
both POS labels and dependency relations receiving highest weights suggest a misleading ``\textit{PERSON}'' label\footnote{``\textit{Bill}'' is part of the ``\textit{PERSON}'' entity in most cases.} because of their context features,
so that an incorrect NER prediction is expected if treating
the three types of syntactic information equally.
%
However, the syntactic constituents give strong indication of the correct label through the word ``\textit{Rights}'' for a ``\textit{LAW}'' entity.
Later,
the syntax attention ensures that the constituent information should be emphasized and the gate mechanism also tends to use syntax for this input with higher weights.
Therefore this case clearly illustrates the contribution of each component in our attentive ensemble of syntactic information.
%
%
%
%
%

\section{Related Work}
\label{relate}

%
Recently, neural models play dominant roles in NER because of their effectiveness in capturing contextual information in the text without requiring to extract manually crafted features 
\cite{DBLP:journals/corr/HuangXY15,DBLP:conf/naacl/LampleBSKD16,DBLP:conf/emnlp/StrubellVBM17,DBLP:conf/acl/ZhangY18,DBLP:conf/naacl/PetersNIGCLZ18,DBLP:conf/coling/YadavB18,DBLP:conf/tlt/CetoliBOS18,DBLP:conf/coling/AkbikBV18,DBLP:conf/naacl/AkbikBV19,DBLP:conf/aaai/ChenLDLZK19,DBLP:conf/naacl/DevlinCLT19,DBLP:conf/naacl/ZhuW19,DBLP:conf/acl/LiuYL19,DBLP:conf/emnlp/BaevskiELZA19,DBLP:journals/corr/abs-1911-04474,DBLP:conf/cikm/XuWHL19,DBLP:journals/access/ZhuHFX20,DBLP:conf/aaai/LuoXZ20}.
%
%
However, to enhance NER, it is straightforward to incorporate more knowledge to it than only modeling from contexts.
%
Therefore,
additional resources such as knowledge base \cite{DBLP:conf/acl/KazamaT08,DBLP:conf/konvens/TkachenkoS12,DBLP:conf/acl/SeylerDCHW18,DBLP:conf/acl/LiuYL19,DBLP:conf/naacl/LiuDS19,DBLP:conf/emnlp/GuiZZPFWH19,DBLP:conf/ijcai/GuiM0ZJH19} and syntactic information 
\cite{DBLP:conf/uai/McCallum03,mohit2005syntax,DBLP:conf/naacl/FinkelM09,DBLP:conf/emnlp/LiDWCM17,DBLP:journals/bioinformatics/LuoYYZWLW18,DBLP:conf/tlt/CetoliBOS18,DBLP:conf/emnlp/JieL19} are applied in previous studies.
Particularly,
\newcite{DBLP:journals/bioinformatics/LuoYYZWLW18} and \newcite{DBLP:journals/bioinformatics/DangLNV18} exploited POS labels and syntactic constituents in their methods and found that the combination of them improves NER performance.
Yet they are limited in regarding such syntactic information as gold references and directly concatenated them to the input embeddings, so that noises are expected to affect NER accordingly.
%
%
Compared with them, our model provides an alternative option to leverage syntactic information with attentive ensemble to encode, weight and select them to help NER, which proves its effectiveness and has the potential to be applied in other similar tasks.

\section{Conclusion}
\label{conclu}

In this paper, we proposed a neural model following the sequence labeling paradigm to enhance NER through attentive ensemble of syntactic information.
Particularly, the attentive ensemble consists of three components in a sequence:
each type of syntactic information is encoded by key-value memory networks,
different information types are then weighted in syntax attention,
and the gate mechanism is finally applied to control the contribution of syntax attention outputs to NER for different contexts.
%
In doing so, different syntactic information are comprehensively and selectively learned to enhance NER, where the experimental results on six benchmark datasets in English and Chinese confirm the validity and effectiveness of our model with state-of-the-art performance obtained.

\section*{Acknowledgments}

Xiang Ao is partially supported by the National Natural Science Foundation of China under Grant No.61976204, U1811461, the Project of Youth Innovation Promotion Association CAS and Beijing Nova Program Z201100006820062.

\bibliography{emnlp2020}
\bibliographystyle{acl_natbib}

\vspace{0.2cm}


\section*{Appendix A: Comparison Between BERT and ZEN on Chinese Datasets}

\begin{table}[h]
    \centering
    \small
    \begin{sc}
    \scalebox{0.95}{
    \begin{tabular}{L{3.6cm}C{0.9cm}C{0.9cm}C{0.9cm}}
        \toprule
        \textbf{Embeddings} & \textbf{ON4c} & \textbf{RE} & \textbf{WE} \\
        \midrule
        BERT + Giga + Tencent   & 80.91 & 96.56 & 69.61 \\
        ZEN + Giga + Tencent    & \textbf{81.18}& \textbf{96.62}& \textbf{69.78} \\ 
        \bottomrule
    \end{tabular}}
    \end{sc}
    \vspace{-0.2cm}
    \caption{
    Experimental results ($F1$ scores) of our models (i.e. the full model with attentive ensemble of all syntactic information), where BERT or ZEN is used as one of the three types of embeddings (the others are Giga and Tencent Embedding) in the embedding layer.
    }
    \label{tab:bert}
    \vskip -1.0em
\end{table}

In the main experiments, we use ZEN \cite{DBLP:journals/corr/zen} rather than BERT \cite{DBLP:conf/naacl/DevlinCLT19} as part of the embeddings of input texts on all Chinese datasets.
The reason is that compared to BERT, ZEN achieves better performance across all Chinese datasets, which is demonstrated in Table \ref{tab:bert}. 
In this experiment, the results ($F1$ scores) show the performance of our best performing model (i.e. the full model with attentive ensemble of all syntactic information) on the test set of all three Chinese datasets.
Specifically, either BERT or ZEN is used as one of the three types of embeddings (the others are Giga and Tencent Embedding).

\section*{Appendix B: Effect of Different Context Encoders}

\begin{table}[h]
    \centering
    \small
    \begin{subtable}[t]{0.48\textwidth}
        \small
        \begin{tabular}{L{1.5cm}C{1.7cm}C{0.8cm}C{0.8cm}C{0.8cm}}
        \toprule
        \textbf{Context} & \textbf{Syntactic} & \multirow{2}{*}{\textbf{ON5e}} & \multirow{2}{*}{\textbf{WN16}} & \multirow{2}{*}{\textbf{WN17}} \\
        \textbf{Encoder} & \textbf{Information}  \\
        \midrule
        \multirow{2}{*}{Bi-LSTM}             &           & 88.56 & 51.16 & 48.11 \\
                            & $\surd$   & 89.64 & 53.39 & 49.56 \\
        \midrule
        \multirow{2}{*}{Transformer}         &           & 88.97 & 52.31 & 48.69 \\
                            & $\surd$   & 89.92 & 54.56 & 50.21 \\
        \midrule
        Adapted- &           & 89.32 & 53.81 & 48.96 \\
        Transformer         & $\surd$   & \textbf{90.32} & \textbf{55.14} & \textbf{50.68} \\
        \bottomrule
        \end{tabular}
        \caption{Performance on all English datasets.}
        \vspace{0.3cm}
    \end{subtable}

    \begin{subtable}[t]{0.48\textwidth}
        \small
        \begin{tabular}{L{1.5cm}C{1.6cm}C{0.8cm}C{0.8cm}C{0.8cm}}
        \toprule
        \textbf{Context} & \textbf{Syntactic} & \multirow{2}{*}{\textbf{ON4c}} & \multirow{2}{*}{\textbf{RE}} & \multirow{2}{*}{\textbf{WE}} \\
        \textbf{Encoder} & \textbf{Information}  \\
        \midrule
        \multirow{2}{*}{Bi-LSTM}             &           & 77.32 & 94.81 & 65.72 \\
                            & $\surd$   & 80.03 & 96.08 & 68.11 \\
        \midrule
        \multirow{2}{*}{Transformer}         &           & 78.18 & 95.26 & 67.16 \\
                            & $\surd$   & 80.46 & 96.31 & 69.24 \\
        \midrule
        Adapted- &           & 79.04 & 95.84 & 67.79 \\
        Transformer         & $\surd$   & \textbf{81.18} & \textbf{96.62} & \textbf{69.78} \\
        \bottomrule
        \end{tabular}
        \caption{Performance on all Chinese datasets.}
    \end{subtable}
    \vspace{-0.2cm}
    \caption{Experimental results ($F1$ scores) of our models with and without applying syntactic information (attentive ensemble of all syntactic information) using different types of context encoders.}
    \label{tab:context}
\end{table}

In the main experiments, we use Adapted-Transformer \cite{DBLP:journals/corr/abs-1911-04474} as the context encoder.
In this experiment, we try other two popular context encoders, i.e., Bi-LSTM and Transformer \cite{DBLP:conf/nips/VaswaniSPUJGKP17}, with attentive ensemble of all syntactic information).
%
%
The results ($F1$) on the test set of all datasets are reported in Table \ref{tab:context}, where the corresponding baselines without using syntactic information are also reported for reference.
From the results, we find that all models with attentive ensemble to incorporate syntactic information consistently outperform their corresponding baselines across all English and Chinese datasets, which demonstrates the robustness of our approach to leverage syntactic information to improve NER.

\section*{Appendix C: Hyper-parameter Settings}

\begin{table}[h]
    \centering
    \small
    \begin{sc}
    \begin{tabular}{L{3.0cm}|L{2.5cm}|R{0.8cm}}
        \toprule
         & \textbf{Values} & \textbf{Best} \\
        \midrule
        Dropout rate            & $0$, $0.1$, $0.2$, $0.3$      & $0.2$ \\
        Learning rate           & $e^{-5}$, $e^{-4}$, $e^{-3}$  & $e^{-4}$  \\
        Batch size              & $8$, $16$, $32$               & $32$ \\
        Number of layers        & $1$, $2$, $4$                 & $2$ \\
        Number of head          & $4$, $8$, $12$                & $12$ \\
        Hidden units            & $64$, $128$, $256$            & $128$ \\
        \bottomrule
    \end{tabular}
    \end{sc}
    \vspace{-0.2cm}
    \caption{
    Values tested for different hyper-parameters and the best one used in our experiments.
    }
    \label{tab:hyper}
    \vskip -1.0em
\end{table}

We try different values for hyper-parameters for our model, presented in Table \ref{tab:hyper}.
The best values for all hyper-parameters are also reported, which are obtained by tuning our model with the given hyper-parameter values on the development set of each dataset.

\section*{Appendix D: The Results of Our Models on the Development Set}

\begin{table}[h]
    \centering
    \small
    \begin{sc}
    \scalebox{0.83}{
    \begin{tabular}{L{0.45cm}C{0.2cm}C{0.2cm}C{0.2cm}C{0.65cm}C{0.7cm}C{0.7cm}C{0.65cm}C{0.6cm}C{0.6cm}}
        \toprule
        \multirow{2}{*}{\textbf{$\mathcal{GM}$}}&\multicolumn{3}{c}{\textbf{Syn.}}&\multirow{2}{*}{\textbf{ON5e}}&\multirow{2}{*}{\textbf{WN16}}&\multirow{2}{*}{\textbf{WN17}}&\multirow{2}{*}{\textbf{ON4c}}&\multirow{2}{*}{\textbf{RE}}&\multirow{2}{*}{\textbf{WE}}\cr
        \cmidrule{2-4}
        &\textbf{P.}&\textbf{C.}&\textbf{D.}\cr
        \midrule
         & $\surd$ & $\surd$ &                  & 86.21         & 55.54         & 49.48         & 77.06 & 96.04 & 67.86 \cr
        $\surd$ & $\surd$ & $\surd$ &           & 86.78         & 56.26         & 49.76         & 77.65 & 96.34 & 68.35 \cr
        \midrule
         & $\surd$ & & $\surd$                  & 86.41         & 56.31         & 49.69         & 77.31 & 96.12 & 67.63 \cr
        $\surd$ & $\surd$ & & $\surd$           & 86.84         & 56.84         & 50.02         & 77.56 & 96.31 & 67.86 \cr
        \midrule
         & & $\surd$ & $\surd$                  & 86.58         & 56.56         & 49.75         & 77.42 & 96.12 & 67.52 \cr
        $\surd$ & & $\surd$ & $\surd$           & 86.92         & 57.26         & 50.18         & 77.74 & 96.39 & 68.14 \cr
        \midrule
         & $\surd$ & $\surd$ & $\surd$          & 86.71         & 56.41         & 50.04         & 77.61 & 96.41 & 68.16 \cr
        $\surd$ & $\surd$ & $\surd$ & $\surd$   & \textbf{87.03}& \textbf{57.38}& \textbf{50.51} & \textbf{78.05} & \textbf{96.46} & \textbf{68.92} \cr
        \bottomrule
    \end{tabular}}
    \end{sc}
    \vspace{-0.2cm}
    \caption{
    $F1$ scores of our models under different configurations on the development set of all datasets.
    ``$\mathcal{GM}$'' is the gate mechanism;
    ``P.'', ``C.'' and ``D.''
    refer to
    POS labels, syntactic constituents and dependency relations, respectively. 
    }
    \label{tab:dev}
    \vskip -1.0em
\end{table}

In Table \ref{tab:dev}, we report the experimental results ($F1$) of our models (i.e., with all types of embeddings and KVMN) under different configurations (using syntax attention on different combinations of syntactic information and whether to use the gate mechanism) 
on development set of all datasets.

\end{document}